\documentclass[
]{ceurart}

\usepackage{listings}
\usepackage{wrapfig}
\usepackage{graphicx}
\usepackage[caption=false]{subfig}

\begin{document}

\copyrightyear{2024}
\copyrightclause{Copyright for this paper by its authors.
 Use permitted under Creative Commons License Attribution 4.0
 International (CC BY 4.0).}

\conference{eCom'24: ACM SIGIR Workshop on eCommerce, July 18, 2024, Washington, DC, USA}

\title{Cooperative Multi-Agent Deep Reinforcement Learning in Content Ranking Optimization}


\author[1]{Zhou Qin}[%
orcid=0000-0002-1641-772X,
email=qinzhouhit@hotmail.com,
url=https://zhouqin.info,
]
\fnmark[1]
\cormark[1]
\address[1]{Amazon.com, Inc. 550 Terry Ave N, Seattle, Washington 98109}

\author[1]{Kai Yuan}
\fnmark[1]

\author[1]{Pratik Lahiri}

\author[1]{Wenyang Liu}
\cortext[1]{Corresponding author.}
\fntext[1]{These authors contributed equally.}






\begin{abstract}
In a typical e-commerce setting, Content Ranking Optimization (CRO) mechanisms are employed to surface content on the search page to fulfill customers' shopping missions. CRO commonly utilizes models such as contextual deep bandits model to independently rank content at different positions, e.g., one optimizer dedicated to organic search results and another to sponsored results. However, this regional optimization approach does not necessarily translate to whole page optimization, e.g., maximizing revenue at the top of the page may inadvertently diminish the revenue of lower positions. In this paper, we propose a reinforcement learning based method for whole page ranking to jointly optimize across all positions by: 1) shifting from position level optimization to whole page level optimization to achieve an overall optimized ranking; 2) applying reinforcement learning to optimize for the cumulative rewards instead of the instant reward. We formulate page level CRO as a cooperative Multi-agent Markov Decision Process , and address it with the novel \textbf{M}ulti-\textbf{A}gent \textbf{D}eep \textbf{D}eterministic \textbf{P}olicy Gradient (MADDPG) model. MADDPG supports a flexible and scalable joint optimization framework by adopting a ``centralized training and decentralized execution'' approach. Extensive experiments demonstrate that MADDPG scales to a 2.5 billion action space in the public Mujoco environment, and outperforms the deep bandits modeling by 25.7\% on the offline CRO data set from a leading e-commerce company. We foresee that this novel multi-agent optimization is applicable to similar joint optimization problems in the field of information retrieval.
\end{abstract}


\maketitle

\section{Introduction}
\label{sec:Introduction}

\begin{wrapfigure}{r}{0.36\textwidth}
\includegraphics[width=0.36\textwidth, keepaspectratio=true]{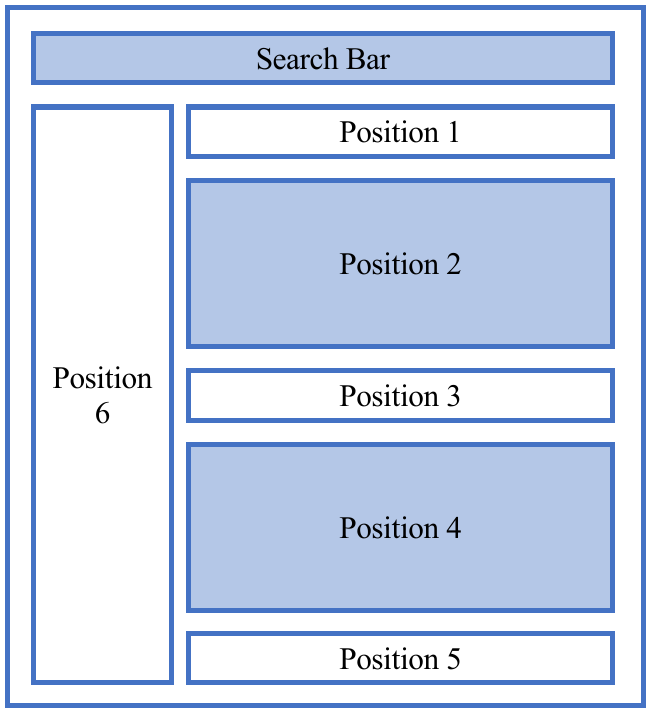}
\caption{Content Positions} \label{fig:page_layout}
\end{wrapfigure}

Traditional e-commerce search results page can be consisted of several positions (a.k.a ``slots''), e.g., shown as the blue boxes in Figure~\ref{fig:page_layout}. Such an explicit design helps maintain a consistent shopping experience for the customers and is easy to maintain. Each position accommodates several rankable pieces of contents. Different positions could also serve different purposes: top-position content (e.g., position 1) includes highly relevant contents or a dedicated module answering or clarifying customers' questions; middle positions can help customers navigate, narrow down, or discover related products and categories; bottom positions may have lower relevance, but aims to inspire customers to either diverge from the original intent or explore their search further. Content providers can schedule their contents to show at particular positions, based on different content properties. When multiple pieces of contents compete for the same position, a content optimizer (a.k.a ranker) is often designed to determine which are the most helpful pieces of contents to surface.  

In common use cases, content ranking optimization (CRO) can be formalized as a ranking problem, such as building contextual deep learning models to optimize for certain objective functions, e.g., revenue, profit, or click-through rate (CTR), etc.~\cite{zhou2019deep} Or more often, a multi-objective optimization problem, i.e., optimizing several metrics simultaneously~\cite{deb2016multi}. In real-world use cases, positions on the search results page are subject to certain restrictions: 1) some positions are designated to fixed content providers; 2) different positions hold contents from different content providers, which are independent from each other; 3) combining multiple positions can lead to a huge action space. These restrictions have resulted in independent ranking decisions for different positions, meaning CRO does not take other positions' information into consideration. 

In this paper, we propose to utilize a multi-agent deep reinforcement learning (RL) solution (MADDPG) to tackle the CRO problem, which aims to 1) achieve a joint whole-page level contextual optimization across multiple positions, by learning the interactions among positions and generating whole page contents combinations; 2) optimize for long term benefits instead of instant benefits; 3) improve customer experience to better fulfill customers' shopping missions. The optimizer of each position is referred as an ``agent'', and contents of each position are referred as ``actions''. All agents act cooperatively to receive collective whole page level rewards. We utilize full RL to optimize for the accumulative future rewards during the entire customer journey, instead of heuristic instant rewards. We believe this novel method is applicable and innovative to other joint optimization applications in e-commerce.

The rest of the paper is organized as follows: Section~\ref{sec:RelatedWorks} describes the related work used in the industry; Section~\ref{sec:Problem} describes the CRO problem formulation; Section~\ref{sec:Modeling} describes the proposed solution; Section~\ref{sec:Evaluation} describes the evaluation and results; and Section 6 summarizes our work and discusses future research directions.

\section{Related Works}
\label{sec:RelatedWorks}

Reinforcement learning has gained significant attention in the field of e-commerce for ranking related problems, where the goal is to present customers with ranked list of items to maximize customer engagement and conversions. Specifically, we focus on the work addressing the high-dimensional action space in the ranking scenario. 

Multi-Variate Testing (MVT)~\cite{hill2017efficient} tackles multi-dimensional joint optimization problem and is widely adopted in e-commerce companies. It builds on top the of multi-armed bandits (MAB) model~\cite{silva2022multi}, and adopts the hill-climbing algorithm~\cite{hill2017efficient} to efficiently explore the high dimensional action space. One concern with applying MVT is the latency of neural networks. Hill-climbing in MVT requires several rounds of inferences. However, CRO typically utilizes the neural networks (NN) based models to estimate the objective. The motivation is to reduce manual feature engineering effort and accommodate advanced features into model, such as embedding. As a result, inference with NN models is expensive. Thus, it is intractable to meet the latency requirement of NN models and MVT. 

Hierarchical RL (HRL)~\cite{takanobu2019aggregating} was proposed to solve the joint optimization problem on e-commerce search page. Optimization is decomposed into 2 sub-tasks: 1) a source selector that decides which sources should be selected at the current page; and 2) an item presenter that decides the presentation order of the items selected from the selected sources in a page to each position. The model fully utilizes the sequential characteristics of user behaviors across different pages to decide both the sources and items to be presented. The item selector is similar to content ranking in CRO, but hierarchical structure is not required in CRO, thus HRL is not practical. However, this approach inspires the joint optimization between page layout optimization (PLO)~\cite{qin2022automate} and content ranking, where PLO decides the page template at a higher-level and CRO decides the contents given the selected page layout at a lower level.

Reinforcement learning has also been applied in different ranking methodologies, e.g., list-wise recommendation~\cite{zhao2017deep} and page-wise recommendation~\cite{zhao2018deep}. The page-wise solution generates items and the strategy to display them on a two-dimension page.

In CRO, we firstly apply MADDPG to the whole page search ranking problem, providing a scalable solution for an industrial setting with tailored modifications.

\section{Problem}
\label{sec:Problem}

\subsection{Combinatorial Optimization}

We formalize the content ranking optimization (CRO) as a combinatorial optimization problem, where the major challenge is to deal with a combinatorial action space. In CRO, an action is content. With more new positions and contents, the action space will grow exponentially. Formally, for a CRO environment with $N$ positions and $n_d$ discrete actions at each position $d$, there will be $\prod_{d=1}^{N}n_d$ possible actions to be considered. Table~\ref{tb:comb_stat} lists the existing contents, containing the possible combinations and the actual max combinations from a single search request: (Actual max is less than the possible number because contents may have different traffic coverage).

\begin{table}[!htbp]\centering
\small
\caption{Content Combinations in CRO}
\begin{tabular}{cccc} \hline 
\multicolumn{4}{c}{Platform: Desktop} \\
\hline
Positions & Region 1 & Region 2 & Region 3 \\
\hline
Position 1     & 50 & 28 & 9 \\
Position 2  & 68 & 52 & 26 \\
Position 3  & 21 & 19 & 5 \\
Position 4  & 18 & 13 & 3 \\
Position 5  & 7 & 8 & 2 \\
Position 6  & 14 & 14 & 4 \\
\hline
Possible combinations  & 125M & 40M & 28,000\\
Actual Max Combinations & 30,720 & 20,400 & 9,600\\
\hline
\end{tabular}
\label{tb:comb_stat}
\end{table}

Such a large action space is intractable for conventional single-agent single-dimensional RL algorithms, such as DQN~\cite{mnih2013playing}, since it is difficult to explore efficiently~\cite{tavakoli2018action}. One solution is to delegate the optimization responsibility to lower-level agents, where each agent optimizes its own dimension, and agents have a way to communicate~\cite{oroojlooy2023review} with other agents to cooperate. As a result, the action space grows linearly for each agent. We formulate CRO as a multi-agent RL due to: 1) optimality: this work evaluates both single-agent RL and multi-agent RL in evaluation Section~\ref{sec:Evaluation}, and demonstrate that multi-agent RL algorithm performs optimally, stably, and scalably in a CRO-like scenario; 2) flexibility: multi-agent RL applies a ``centralized training and decentralized execution'' approach~\cite{oroojlooy2023review}, where each agent is not required to access another agent's observation at execution time. This supports the flexible system structure enabling joint optimization across different applications. For instance, we can deploy trained agents to different services for online execution.

\subsection{Multi-agent MDP in CRO}
In this part, we formalize the problem in CRO using MADDPG. We consider a multi-agent extension of a MDP called Markov games~\cite{oroojlooy2023review} with the tuple $G=(N, S, \mathcal{A}, R, P,  \mathcal{O}, \gamma)$. Here, $N$ is the number of agents; $\mathcal{A} = \{\mathcal{A}_1, \mathcal{A}_2, ..., \mathcal{A}_N\}$ is the discrete set of actions for the agents; $S$ is the state space; $R$
is the reward function; and $\mathcal{O}=\{\mathcal{O}_1, \mathcal{O}_2, ..., \mathcal{O}_N\}$ is the set of observations for each agent. To choose actions, each
agent $i$ uses a stochastic policy $\pi_{\theta}: \mathcal{O}_{i} \times \mathcal{A}_{i} \rightarrow [0, 1]$, which produces the next state according to the state transition
function $P:S \times \mathcal{A}_{1} \times \mathcal{A}_{2} \times ... \times \mathcal{A}_{N} \rightarrow S$. Each agent i obtains rewards as a function of the state and the agent’s action
$r_{i}: S \times \mathcal{A}_{i} \rightarrow \mathbb{R}$, and receives a local observation correlated with the state $o_{i}: S \rightarrow \mathcal{O}_{i}$. The goal of each agent $i$ is to learn
a policy $\pi_{i}:S \rightarrow \mathcal{A}$ which maximizes its own total expected return $R_{i} =  \sum_{i=0}^{T} \gamma^{t}r_{i}^{t}$, where $\gamma$ is a discount factor and $T$ is the time horizon. Notably, since CRO is a fully cooperative environment, at each time step $t$, all agents observe the same joint reward $r_{1}^t = r_{2}^t =...=r_{N}^{t}=r^{t} $. Concept mapping in CRO will be like $r1 * 0.99^0 + r2 * 0.99^1 + r3 * 0.99^3 + ...$.

\begin{itemize}
    \item $Environment$: e-commerce website and customers;
    \item $Episode$: a customer session on the same day;
    \item $S$: a customer’s shopping state which represents the global context $x$, such as region, query, device, membership status; 
    \item $Terminal State$: when the session is abandoned (no more interactions) at the end of the day;
    \item $P$ : state transition probability;
    \item $N$ : position agents to choose content at their respective positions, such as top/middle/bottom/side positions;
    \item $\mathcal{A}_{i}$: content candidates at position $i$;
    \item $\mathcal{O}_{i}$: the observation of position $i$, consisting of two parts:
        \subitem global observation $x$;
        \subitem local observation  $\mathcal{o}_{i}$ regarding the content features at position $i$, such as CTRs, ads bids, etc.;
    \item $R$: the reward which CRO receives from the environment, detailed discussion at 4.2 section;
\end{itemize}

At each time step, a customer issues a request to CRO, CRO observes the customer's shopping state, chooses a content combination and display to the customer, and then receives the rewards from the session. CRO goal is to learn the policy $\pi_{i}$ for each agent $i$ which maximizes the expected return across all sessions.

\subsection{Mix-offline RL System in CRO}

\begin{figure*}[h]
  \centering
\includegraphics[width=0.7\textwidth,keepaspectratio=true]{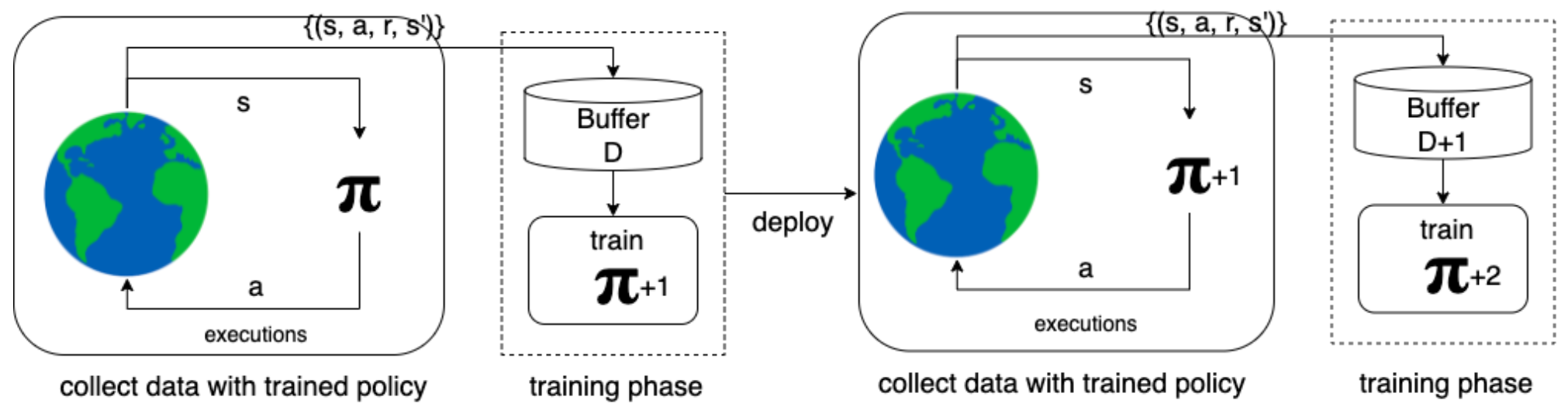}
  \caption{Mix-Offline RL System in CRO}
  \label{fig:mix_offline}
\end{figure*}

It is intractable to collect business rewards instantly online, which is the key limitation preventing online RL. As Figure~\ref{fig:mix_offline} shows, CRO is defined as a mix-offline RL system in the current scenario: every day, a trained agent is deployed online to interact with the environment; at the end of the day, the trajectories are collected to train the agent, then the trained agent is deployed online to act. CRO is closer to an offline RL scenario where the agent can interact with the environment.

\section{Modeling}
\label{sec:Modeling}

\subsection{MADDPG}
We utilize the MADDPG model~\cite{lowe2017multi} to solve the CRO problem. MADDPG is a multi-agent version of actor-critic method (MAAC)~\cite{konda1999actor}. It adopts the framework of centralized training with decentralized execution approach~\cite{oroojlooy2023review}. A key characteristic of MADDPG is utilizing a fully observable critic which involves the observations and actions of all agents to ease joint training. Each agent trains a Deep Deterministic Policy Gradient~\cite{lillicrap2015continuous} model, which concurrently learns a Q-function and a policy. The actor $\pi_{\theta_{i}}(o_i)$ with policy weights $\theta_{i}$ observes the local observations $o_i$, while the critic $Q_{i}^{\mu}$ is allowed to access the observations, actions, and the target policies of all agents during training. The critic of each agent concatenates all state-actions together as the input and uses the local reward to obtain the corresponding Q-value. Formally, consider an environment with $N$ agents and policies parameterized by $\theta = \{\theta_{1}, ..., \theta_{N}\}$, and $\pi=\{\pi_1,...,\pi_{N}\}$ as the set of agent policies. The gradient of the expected return for agent i, $J(\theta_{i}) = \mathbb{E} [R_{i}]$ is:

\begin{equation}
\label{eq:eq1}
    \nabla_{\theta_{i}} J(\theta_{i}) = \mathbb{E}_{s \sim p^{\pi}, a \sim \pi^{\theta}} \Biggl[\nabla_{\theta_{i}}\log \pi_{i} (a_{i}|o_{i})Q_{i}^{\pi}(x, a_{1}, ..., a_{N})\Biggl]
\end{equation}

where $Q^{\pi}(x, a_{1}, ..., a_{N})$ is a centralized action-value function that takes the actions of all agents, $a_{1}, ..., a_{N}$ as input, and the concatenated state information $x=(o_{1}, ..., o_{N})$, and outputs the Q-value for agent $i$. We then extend the work with deterministic policies $\mu_{i}$ with parameters $\theta_{i}$, so the gradient can be written as:

\begin{equation}
\label{eq:eq2}
    \nabla_{\theta_{i}} J (\mu_{i}) = \mathbb{E}_{x, a \sim D} \Biggl[ \nabla_{\theta_{i}} \mu_{i} (a_{i}|o_{i}) \nabla_{a_{i}} Q_{i}^{\mu}(x, a_{i}, ..., a_{N})|_{a_{i}=\mu_{i}(o_i)})\Biggl ]
\end{equation}

Here the experience replay buffer $\mathcal{D}$ contains the tuples $(x, x^{'}, a_{1}, \\..., a_{N}, r)$, recording the experience of all agents. The centralized action-value function $Q_{i}^{\mu}$ is updated as follows:

\begin{equation}
\label{eq:eq3}
    L(\theta_{i})=\mathbb{E}_{x,a,r,x^{'}} \Biggl[ (Q_{i}^{\mu} (x, a_{i}, ..., a_{N}) - (r + \gamma Q_{i}^{\mu^{'}} (x^{'}, a_{1}^{'}, ..., a_{N}^{'})) |_{a_{j}^{'}=\mu_{j}^{'}(o_{j}} )^{2} \Biggl]
\end{equation}

where $\mu^{'}=\{\mu_{\theta_{1}^{'}}, ..., \mu_{\theta_{N}^{'}}\}$ is the set of target policies with delayed parameters $\theta_{i}^{'}$. This method resolves the non-stationarity issue~\cite{kobayashi2021t}, as $P(s^{'}|s, a_{1}, ..., a_{N}, \pi_{1}, ..., \pi_{N})=P(s^{'}|s,a_{1},...,a_{N},\pi_{1}^{'},...,\pi_{N}^{'})$ for any $\pi_{i} \neq \pi_{i}^{'}$, since the environment returns the same next-state regardless of the changes in the policy of other agents.

\subsection{Reward}

Rewards are critical to an RL system. In our setting, we formulate the reward as follows:

\begin{equation}
\label{eq:reward}              
    R_t=Revenue_t+Profit_t+\alpha*LongTerm_t + \beta*Clicks_t + \\ \gamma*Abandonments_t
\end{equation}

\begin{figure}[htbp]
  \centering
\includegraphics[width=0.7\textwidth,keepaspectratio=true]{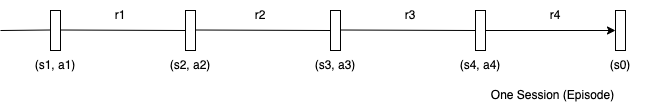}
  \caption{Rewards of One Episode}
  \label{fig:rewards}
\end{figure}

Clicks and Abandonments are part of the reward due to two reasons: 1) we aim to encourage more engagements in the session and reduce session abandonments, which are an unbiased indicators and indicate long-term value; 2) to enable rewards shaping~\cite{ng1999policy} to stabilize the learning since purchases are sparse. Figure~\ref{fig:rewards} shows how to collect the rewards from the session. $s$ represents the customer state, $a$ represents the content combination displayed to the customer, and reward $r_t$ is the total reward received between impressions (time steps). $s_0$ is the terminal state. From this episode, 4 transitions are collected: $(s_1, a_1, r_1, s_2, 0), (s_2, a_2, r_2, s_3, 0), (s_3, a_3, r_3, s_4, 0), (s_4, a_4, r_4, s_0, 1)$.

\subsection{Exploration}
We sample actions from the Gumbel Softmax~\cite{jang2016categorical} action distributions with a temperature of 1.0 for each agent. This is the conventional noisy-based method in RL. With higher temperature, the model encourages more explorations. For new and unrecognized contents, we will assign a fixed probability to ensure they can be explored.

\section{Evaluation}
\label{sec:Evaluation}

We conduct extensive evaluations both in a public Mujoco environment~\cite{todorov2012mujoco} and on a CRO offline data set.

\subsection{Evaluation On HalfCheetah-v2}
\subsubsection{Environment}

We choose the Mujoco~\cite{todorov2012mujoco} environment HalfCheetah-v2~\cite{halfcheetah} to simulate the CRO scenario. The robot in this environment has six limbs, with each limb representing one continuous action $a_i \in [0.0, 1.0]$. The goal is to train agents to control these limbs cooperatively to move faster and receive higher rewards. Since CRO has a discrete action space - contents, we need to discretize the continuous space into a discrete space to bridge the gap. For instance, when the per-dimension action size is set to 5, we discretize the action space to be [1.0, 0.5, 0.0, 0.5, 1.0] and at every time step, the agents interact with the environment using one of those discretized values. If the max per-dimension action size is 5, the action space expands to $5^6 = 15625$; if action size is 50, the action space reaches $50^6 = 15.6B$, which simulates a extremely complex cooperative scenario.

\subsubsection{Benchmarks}
We leverage the following models as benchmarks:
\begin{itemize}
    \item \textbf{DQN}: The classical DQN model. We flatten multi-dimensional action spaces into a single dimension
    \item \textbf{CQL}: Conservative Q-learning~\cite{kumar2020conservative}, tweaked DQN model with a conservative Q function. CQL is designed to address the over-estimation issue~\cite{kumar2020conservative} in an offline RL setting using an off-policy algorithm. We also flatten action spaces into a single dimension;
    \item \textbf{BranchingDQN}: The action branching architecture~\cite{tavakoli2018action} was proposed to address the large action space problem in single-agent RL. The key insight is that to solve problem in combinatorial action space, it is possible to optimize for each action dimension with a degree of independence. The shared network module computes a latent representation of the input state that is then propagated to the several action branches. Each branch is responsible for controlling an individual degree of freedom, and the concatenation of the selected sub-actions results in a joint-action tuple;
    \item \textbf{BranchingCQL}: A conservative version of BranchingDQN to fit into an offline RL scenario;
    \item \textbf{MADDPG}: This paper's proposed model.
    
\end{itemize}

\subsubsection{Online Setting}
We run models in HalfCheetah-v2 in an online fashion with the discrete per-dimension action sizes of [5, 10, 25, 50] resulting in [15K, 1M, 244M, 15.6B] action spaces. Models are run 3 times with 1,000 episodes and mean rewards are averaged. The CQL and DQN model cannot be tested with action size > 10, due to Out-of-Memory issues on the GPU. All policy and critic networks have the same 2 MLP layers with 256 hidden units and ReLU activation function~\cite{agarap2018deep}. Target networks are updated by the $\tau$-soft update method~\cite{konda1999actor} where $\tau = 0.001$. For training the CQL, DQN, and Branching*, we utilize an epsilon-greedy~\cite{dann2022guarantees} strategy as the behavioral policy~\cite{fujimoto2019off} with an exponential decay $\epsilon = (0.01 + 0.99 * e^{\frac{-steps}{1e5}})$; for MADDPG, we sample actions from a Gumbel Softmax~\cite{jang2016categorical} action distribution. The experience replay buffer is set to 105, and the batch size is set to 128 for all models.

\begin{figure}[htbp]
  \centering
  
\subfloat[Mean Reward, Action Size 5]{
\includegraphics[width=0.45\textwidth,keepaspectratio=true]{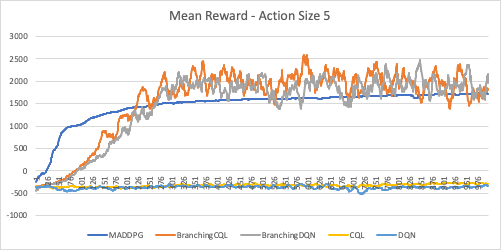}}
\vspace*{-3pt}
\subfloat[Mean Reward, Action Size 10]{
\includegraphics[width=0.45\textwidth,keepaspectratio=true]{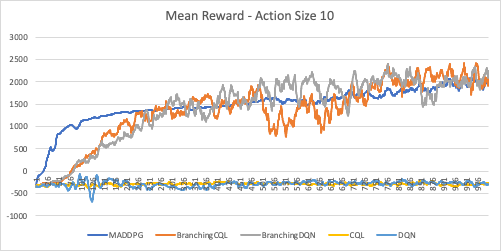}}
\vspace*{-3pt}
\subfloat[Mean Reward, Action Size 25]{
\includegraphics[width=0.45\textwidth,keepaspectratio=true]{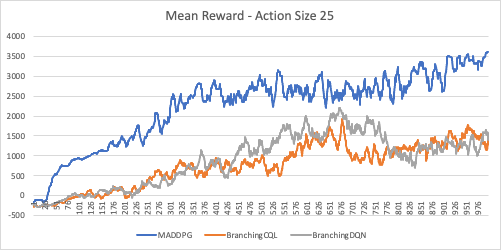}}
\vspace*{-3pt}
\subfloat[Mean Reward, Action Size 50]{
\includegraphics[width=0.45\textwidth,keepaspectratio=true]{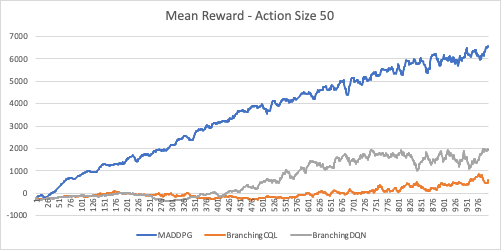}}

  \caption{Model Performance in Online Setting}
  \label{fig:online_eval}
\end{figure}

The results in Figure~\ref{fig:online_eval} indicates that typical DQN and CQL models are intractable for handling the large action space due to the insufficient explorations, which is expected. Branching models learn well at relatively low dimension action size but do not scale as well as MADDPG. MADDPG demonstrates consistent and optimal performance in all online experiments.

\subsubsection{Offline Setting}
The online study demonstrates that MADDPG scales well in an online setting. However, CRO is more suitable for an offline setting. Specifically, the CRO agent is required to learn from the trajectories collected by online policies at the end of the day. In this section, we conduct experiments to verify whether MADDPG can perform well in an offline setting. We still apply per-dimension action sizes of [5, 10, 25, 50] to observe its scalability. Offline learning consists of two steps: 1) collecting trajectories from some online policies (or random policy); 2) training the offline agent with these trajectories. To generate the trajectories, we adopt the D4RL approach~\cite{fu2020d4rl}: first, we train an online MADDGP agent in a HalfCheetah-v2 environment, recording its mean rewards and saving its checkpoints until the model converges; second, we load 3 agents: 2 agents from checkpoints with 30\% and 100\% performance, and 1 random agent; then, we have the 3 agents interact with the environment (not train) to collect 3M trajectories equally (1M - 100\% - expert, 1M - 30\% - medium, 1M - random); finally, we train the models on this data set and observe the performance trending during training. All model settings are identical to the online setting.

\begin{figure}[htbp]
  \centering
  
\subfloat[Mean Reward, Action Size 5]{
\includegraphics[width=0.45\textwidth,keepaspectratio=true]{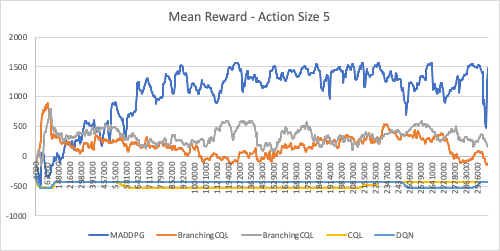}}
\vspace*{-3pt}
\subfloat[Mean Reward, Action Size 5]{
\includegraphics[width=0.45\textwidth,keepaspectratio=true]{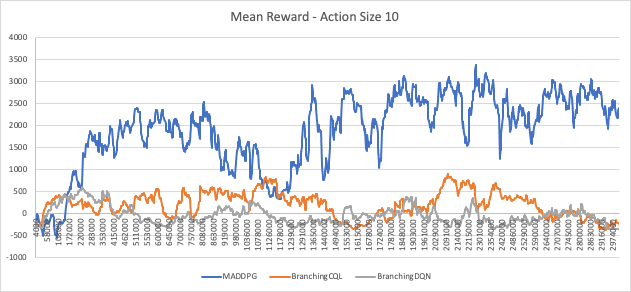}}
\vspace*{-3pt}
\subfloat[Mean Reward, Action Size 5]{
\includegraphics[width=0.45\textwidth,keepaspectratio=true]{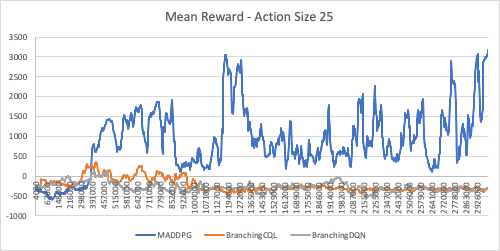}}
\vspace*{-3pt}
\subfloat[Mean Reward, Action Size 5]{
\includegraphics[width=0.45\textwidth,keepaspectratio=true]{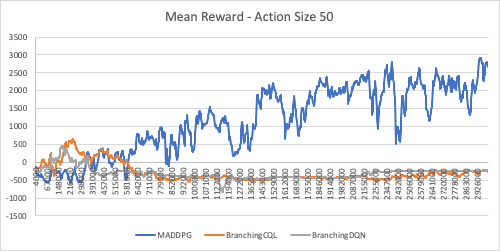}}

  \caption{Model Performance in Offline Setting}
  \label{fig:offline_eval}
\end{figure}

The results in Figure~\ref{fig:offline_eval} demonstrates that 1) conservative models can learn better than DQN in an offline setting, which aligns with their design purpose, but CQL does not scale well with larger action sizes; 2) MADDPG scales well in the offline setting.

\subsection{Evaluation on CRO Data Set}
In this section, we conduct experiments to evaluate MADDPG on the CRO data set. For simplicity, we extract a business metric as the reward for benchmarking. We evaluate the models on desktop traffic only and consider joint optimization for 3 positions - top/middle/bottom positions, as they have the most impacts and content competition.

\subsubsection{Trajectory Preparation}
Trajectory data serves as the training source for RL. We follow a standard procedure to extract the training trajectories from CRO logs. As a result, we extracted 40M training trajectories from desktop traffic. The trajectories are from baseline policies: 5\% random policy (generated by 5\% epsilon greedy) and 95\% deep bandits policy. For evaluation trajectories, we follow the same steps but use an exploration filter to extract 4M trajectories from the random policy.

\subsubsection{Benchmarks}
We leverage the following models and and rewards as benchmarks:
\begin{itemize}
    \item \textbf{Random}: It randomly selects content combinations from available options.
    \item \textbf{Baseline (Multi-component reward)}: It is a baseline ranker with click-based Multi-objective Optimization objective.
    \item \textbf{Position-level bandits}: It is identical to the baseline model structure, past search queries are tokenized by the pre-trained SentencePiece tokenizer~\cite{kudo2018sentencepiece} and encoded by Embedding and LSTM~\cite{hochreiter1997long} layers. Then we concatenate embedding with other inputs and feed it to an MLP layer with 256 hidden units. The final output is one unit with a Mean Squared Error (MSE) loss. This model is similar to the baseline, with the major differences in the features and reward.
    \item \textbf{Page-level bandits}: It utilizes an MLP structure with a multi-head output structure. Each "head" represents one position and outputs the content probability for available contents at this position. Features are the same as described above: past search queries, customer context, top position content ID, inline stop content ID, and bottom position content IDs.
    \item \textbf{MADDPG w/ gamma=0}: We train it on trajectories with a discounted factor=0.0. State is encoded by SentencePiece, Embedding and LSTM layers as well. Since the discounted factor is 0, this actually acts as a bandits model which optimizes for the instant reward without considering the future rewards.
    \item \textbf{MADDPG w/ gamma=0.99}: We train it on trajectories with a discount factor as 0.99; others settings are the same as described above.
\end{itemize}

The difference between the baseline and random models demonstrates the impact of the baseline model. The difference between position-level and page-level bandits presents the benefit of joint optimization. The difference between ``MADDPG w/ gamma=0.99'' and ``MADDPG w/ gamma=0'' illustrates the benefit of RL over bandits.

\subsubsection{Metrics}
We adopt the Inverse Propensity Scoring (IPS) method~\cite{williamson2014introduction} to evaluate the trained policies: Given logged exploration data with a policy $\mu$, and an evaluation policy $\pi$, the following IPS estimator of $\pi$ is unbiased:

\begin{equation}
    \hat{V}_{IPS} (\pi) = \frac{1}{n} \sum_{i=1}^{n} \frac{\pi(a_i|x_i)}{\mu(a_i|x_i)} r_{i}
\end{equation}

where $n$ is the trajectory count, $r_i$ is the trajectory reward. Since it is a random policy, $\mu(a_i|x_i) = \frac{1}{|a|}$ is extracted by softmax [11] from the output of models. The higher the IPS, the better the policy. Due to the high variance of the reward and IPS calculation, we choose 3 scores for the benchmark: 1) IPS on source rewards; 2) IPS on logged rewards; 3) IPS on indicator rewards (1 when source reward > 0 otherwise 0). The results are shown in Table 2.

\begin{table*}[!htbp]\centering
\small
\caption{content Combinations in CRO}
\begin{tabular}{cccc} \hline 
\textbf{Model} & \textbf{IPS} & \textbf{IPS (logged)} & \textbf{IPS (indicator)} \\
\hline
Random     & 0.92649 (0.91583, 0.93713) & 0.10628 (0.10586, 0.10670) & 0.15863 (0.15780, 0.15946) \\
Baseline  & 0.92887 (0.91815, 0.93958) & 0.10639 (0.10597, 0.10681) & 0.15895 (0.15811, 0.15978) \\
Position-Level Bandits  & 0.93238 (0.91965, 0.9451) & 0.10679 (0.10632, 0.10726) & 0.15970 (0.15874, 0.16067) \\
Page-Level Bandits  & 1.06876 (1.04743, 1.09008) & 0.12579 (0.12486, 0.12672) & 0.18569 (0.18383, 0.18754) \\
MADDPG gamma=0.0  & 1.12623 (1.11051, 1.14193) & 0.12755 (0.12692, 0.12819) & 0.19239 (0.19108, 0.19369) \\
MADDPG gamma=0.99  & 1.1684 (1.15286, 1.18394) & 0.13353 (0.13288, 0.134188) & 0.20086 (0.19954, 0.20218) \\
\hline
\end{tabular}
\label{tb:cl_cal_stat}
\end{table*}

\begin{figure}[htbp]
  \centering
\includegraphics[width=0.55\textwidth,keepaspectratio=true]{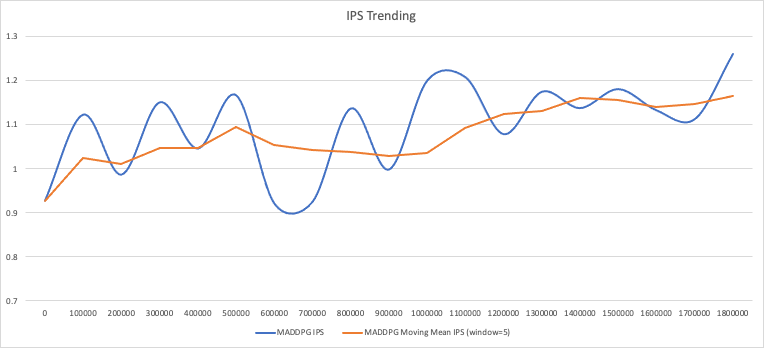}
  \caption{IPS Trend}
  \label{fig:ips_training}
\end{figure}

Table 2 demonstrates that MADDPG outperforms the the bandits model with logged and unlogged reward. The difference between ``MADDPG with gamma=0.99'' and ``MADDPG with gamma=0'' shows that RL is beneficial by considering state transition and delayed reward. The difference between page-level and position-level models shows that ranking holistically is beneficial. The baseline IPS is not high, which we assume is due to the composite effect of different objectives and the over-estimation issue. We also conducted an experiment to verify whether MADDPG can gradually learn during training. We periodically dump the model during training and evaluate it on the CRO evaluation data set and observe the IPS trending. Figure~\ref{fig:ips_training} shows the IPS trending, demonstrating that MADDPG can learn during offline model training.

\subsection{Online A/B Test Performance}
We implemented this approach on one of the leading e-commerce platforms and ran an A/B test to verify its performance. We tested with both discount factor as 0 (refer to as Treatment 1) and 0.99 (refer to as Treatment 2), optimizing a multi-component reward function. By training and deploying it on the online traffic, both treatments demonstrated incremental multi-million revenue value. However, treatment 2 resulted in significant profit losses, leading to a flat overall reward, i.e., a larger discount factor resulted in poorer performance in our setting.

\subsection{Evaluation Summary}

The evaluation demonstrates that the MADDPG model scales
well in both online and offline multi-agent environment
setting, and is able to learn from the CRO offline data set.
MADDPG also significantly outperforms the baseline
like bandits model by 25.7\% on IPS evaluation. We also evaluated this through online A/B test and the proposed solution brought incremental multi-million revenue.

\section{Conclusion}
\label{sec:Conclusion}

In this paper, we frame page-level content ranking as a joint optimization problem, and tackle it using the novel MADDPG model. We conduct extensive experiments on both online Mujoco environment and an offline CRO data set. The evaluations demonstrate that MADDPG scales well up to a 2.5 billion action space and outperforms the baseline bandits model by 25.7\%. In the future, we will further invest in multi-agent RL in several directions: 1) explainability: interpretation of why an agent takes certain action is important for content owners but this is not a trivial task, especially for neural network-based RL. Fortunately we have several directions to explore, such as reward decomposition~\cite{juozapaitis2019explainable} and shapley on Q-values~\cite{wang2020shapley}; 2) efficient exploration: exploration is critical for RL. This paper applies a noisy-based method which may not be the optimal for new contents. We will test entropy-based and variance-based exploration methods; 3) enhanced lifelong RL: retraining new models with a saved experience replay when encountering new contents is not ideal. We will research transfer RL to support the onboarding of new contents without needing to retrain the models.


\bibliographystyle{ACM-Reference-Format}
\bibliography{8bibliography}

\begin{thebibliography}{29}
\expandafter\ifx\csname natexlab\endcsname\relax\def\natexlab#1{#1}\fi
\providecommand{\url}[1]{\texttt{#1}}
\providecommand{\href}[2]{#2}
\providecommand{\path}[1]{#1}
\providecommand{\DOIprefix}{doi:}
\providecommand{\ArXivprefix}{arXiv:}
\providecommand{\URLprefix}{URL: }
\providecommand{\Pubmedprefix}{pmid:}
\providecommand{\doi}[1]{\href{http://dx.doi.org/#1}{\path{#1}}}
\providecommand{\Pubmed}[1]{\href{pmid:#1}{\path{#1}}}
\providecommand{\bibinfo}[2]{#2}
\ifx\xfnm\relax \def\xfnm[#1]{\unskip,\space#1}\fi
\bibitem[{Zhou et~al.(2019)Zhou, Mou, Fan, Pi, Bian, Zhou, Zhu, and
  Gai}]{zhou2019deep}
\bibinfo{author}{G.~Zhou}, \bibinfo{author}{N.~Mou}, \bibinfo{author}{Y.~Fan},
  \bibinfo{author}{Q.~Pi}, \bibinfo{author}{W.~Bian},
  \bibinfo{author}{C.~Zhou}, \bibinfo{author}{X.~Zhu},
  \bibinfo{author}{K.~Gai},
\newblock \bibinfo{title}{Deep interest evolution network for click-through
  rate prediction},
\newblock in: \bibinfo{booktitle}{Proceedings of the AAAI conference on
  artificial intelligence}, volume~\bibinfo{volume}{33}, \bibinfo{year}{2019},
  pp. \bibinfo{pages}{5941--5948}.
\bibitem[{Deb et~al.(2016)Deb, Sindhya, and Hakanen}]{deb2016multi}
\bibinfo{author}{K.~Deb}, \bibinfo{author}{K.~Sindhya},
  \bibinfo{author}{J.~Hakanen},
\newblock \bibinfo{title}{Multi-objective optimization},
\newblock in: \bibinfo{booktitle}{Decision sciences}, \bibinfo{publisher}{CRC
  Press}, \bibinfo{year}{2016}, pp. \bibinfo{pages}{161--200}.
\bibitem[{Hill et~al.(2017)Hill, Nassif, Liu, Iyer, and
  Vishwanathan}]{hill2017efficient}
\bibinfo{author}{D.~N. Hill}, \bibinfo{author}{H.~Nassif},
  \bibinfo{author}{Y.~Liu}, \bibinfo{author}{A.~Iyer},
  \bibinfo{author}{S.~Vishwanathan},
\newblock \bibinfo{title}{An efficient bandit algorithm for realtime
  multivariate optimization},
\newblock in: \bibinfo{booktitle}{Proceedings of the 23rd ACM SIGKDD
  International Conference on Knowledge Discovery and Data Mining},
  \bibinfo{year}{2017}, pp. \bibinfo{pages}{1813--1821}.
\bibitem[{Silva et~al.(2022)Silva, Werneck, Silva, Pereira, and
  Rocha}]{silva2022multi}
\bibinfo{author}{N.~Silva}, \bibinfo{author}{H.~Werneck},
  \bibinfo{author}{T.~Silva}, \bibinfo{author}{A.~C. Pereira},
  \bibinfo{author}{L.~Rocha},
\newblock \bibinfo{title}{Multi-armed bandits in recommendation systems: A
  survey of the state-of-the-art and future directions},
\newblock \bibinfo{journal}{Expert Systems with Applications}
  \bibinfo{volume}{197} (\bibinfo{year}{2022}) \bibinfo{pages}{116669}.
\bibitem[{Takanobu et~al.(2019)Takanobu, Zhuang, Huang, Feng, Tang, and
  Zheng}]{takanobu2019aggregating}
\bibinfo{author}{R.~Takanobu}, \bibinfo{author}{T.~Zhuang},
  \bibinfo{author}{M.~Huang}, \bibinfo{author}{J.~Feng},
  \bibinfo{author}{H.~Tang}, \bibinfo{author}{B.~Zheng},
\newblock \bibinfo{title}{Aggregating e-commerce search results from
  heterogeneous sources via hierarchical reinforcement learning},
\newblock in: \bibinfo{booktitle}{The World Wide Web Conference},
  \bibinfo{year}{2019}, pp. \bibinfo{pages}{1771--1781}.
\bibitem[{Qin and Liu(2022)}]{qin2022automate}
\bibinfo{author}{Z.~Qin}, \bibinfo{author}{W.~Liu},
\newblock \bibinfo{title}{Automate page layout optimization: An offline deep
  q-learning approach},
\newblock in: \bibinfo{booktitle}{Proceedings of the 16th ACM Conference on
  Recommender Systems}, \bibinfo{year}{2022}, pp. \bibinfo{pages}{522--524}.
\bibitem[{Zhao et~al.(2017)Zhao, Zhang, Xia, Ding, Yin, and
  Tang}]{zhao2017deep}
\bibinfo{author}{X.~Zhao}, \bibinfo{author}{L.~Zhang},
  \bibinfo{author}{L.~Xia}, \bibinfo{author}{Z.~Ding},
  \bibinfo{author}{D.~Yin}, \bibinfo{author}{J.~Tang},
\newblock \bibinfo{title}{Deep reinforcement learning for list-wise
  recommendations},
\newblock \bibinfo{journal}{arXiv preprint arXiv:1801.00209}
  (\bibinfo{year}{2017}).
\bibitem[{Zhao et~al.(2018)Zhao, Xia, Zhang, Ding, Yin, and
  Tang}]{zhao2018deep}
\bibinfo{author}{X.~Zhao}, \bibinfo{author}{L.~Xia},
  \bibinfo{author}{L.~Zhang}, \bibinfo{author}{Z.~Ding},
  \bibinfo{author}{D.~Yin}, \bibinfo{author}{J.~Tang},
\newblock \bibinfo{title}{Deep reinforcement learning for page-wise
  recommendations},
\newblock in: \bibinfo{booktitle}{Proceedings of the 12th ACM conference on
  recommender systems}, \bibinfo{year}{2018}, pp. \bibinfo{pages}{95--103}.
\bibitem[{Mnih et~al.(2013)Mnih, Kavukcuoglu, Silver, Graves, Antonoglou,
  Wierstra, and Riedmiller}]{mnih2013playing}
\bibinfo{author}{V.~Mnih}, \bibinfo{author}{K.~Kavukcuoglu},
  \bibinfo{author}{D.~Silver}, \bibinfo{author}{A.~Graves},
  \bibinfo{author}{I.~Antonoglou}, \bibinfo{author}{D.~Wierstra},
  \bibinfo{author}{M.~Riedmiller},
\newblock \bibinfo{title}{Playing atari with deep reinforcement learning},
\newblock \bibinfo{journal}{arXiv preprint arXiv:1312.5602}
  (\bibinfo{year}{2013}).
\bibitem[{Tavakoli et~al.(2018)Tavakoli, Pardo, and
  Kormushev}]{tavakoli2018action}
\bibinfo{author}{A.~Tavakoli}, \bibinfo{author}{F.~Pardo},
  \bibinfo{author}{P.~Kormushev},
\newblock \bibinfo{title}{Action branching architectures for deep reinforcement
  learning},
\newblock in: \bibinfo{booktitle}{Proceedings of the aaai conference on
  artificial intelligence}, volume~\bibinfo{volume}{32}, \bibinfo{year}{2018}.
\bibitem[{Oroojlooy and Hajinezhad(2023)}]{oroojlooy2023review}
\bibinfo{author}{A.~Oroojlooy}, \bibinfo{author}{D.~Hajinezhad},
\newblock \bibinfo{title}{A review of cooperative multi-agent deep
  reinforcement learning},
\newblock \bibinfo{journal}{Applied Intelligence} \bibinfo{volume}{53}
  (\bibinfo{year}{2023}) \bibinfo{pages}{13677--13722}.
\bibitem[{Lowe et~al.(2017)Lowe, Wu, Tamar, Harb, Pieter~Abbeel, and
  Mordatch}]{lowe2017multi}
\bibinfo{author}{R.~Lowe}, \bibinfo{author}{Y.~I. Wu},
  \bibinfo{author}{A.~Tamar}, \bibinfo{author}{J.~Harb},
  \bibinfo{author}{O.~Pieter~Abbeel}, \bibinfo{author}{I.~Mordatch},
\newblock \bibinfo{title}{Multi-agent actor-critic for mixed
  cooperative-competitive environments},
\newblock \bibinfo{journal}{Advances in neural information processing systems}
  \bibinfo{volume}{30} (\bibinfo{year}{2017}).
\bibitem[{Konda and Tsitsiklis(1999)}]{konda1999actor}
\bibinfo{author}{V.~Konda}, \bibinfo{author}{J.~Tsitsiklis},
\newblock \bibinfo{title}{Actor-critic algorithms},
\newblock \bibinfo{journal}{Advances in neural information processing systems}
  \bibinfo{volume}{12} (\bibinfo{year}{1999}).
\bibitem[{Lillicrap et~al.(2015)Lillicrap, Hunt, Pritzel, Heess, Erez, Tassa,
  Silver, and Wierstra}]{lillicrap2015continuous}
\bibinfo{author}{T.~P. Lillicrap}, \bibinfo{author}{J.~J. Hunt},
  \bibinfo{author}{A.~Pritzel}, \bibinfo{author}{N.~Heess},
  \bibinfo{author}{T.~Erez}, \bibinfo{author}{Y.~Tassa},
  \bibinfo{author}{D.~Silver}, \bibinfo{author}{D.~Wierstra},
\newblock \bibinfo{title}{Continuous control with deep reinforcement learning},
\newblock \bibinfo{journal}{arXiv preprint arXiv:1509.02971}
  (\bibinfo{year}{2015}).
\bibitem[{Kobayashi and Ilboudo(2021)}]{kobayashi2021t}
\bibinfo{author}{T.~Kobayashi}, \bibinfo{author}{W.~E.~L. Ilboudo},
\newblock \bibinfo{title}{T-soft update of target network for deep
  reinforcement learning},
\newblock \bibinfo{journal}{Neural Networks} \bibinfo{volume}{136}
  (\bibinfo{year}{2021}) \bibinfo{pages}{63--71}.
\bibitem[{Ng et~al.(1999)Ng, Harada, and Russell}]{ng1999policy}
\bibinfo{author}{A.~Y. Ng}, \bibinfo{author}{D.~Harada},
  \bibinfo{author}{S.~Russell},
\newblock \bibinfo{title}{Policy invariance under reward transformations:
  Theory and application to reward shaping},
\newblock in: \bibinfo{booktitle}{Icml}, volume~\bibinfo{volume}{99},
  \bibinfo{organization}{Citeseer}, \bibinfo{year}{1999}, pp.
  \bibinfo{pages}{278--287}.
\bibitem[{Jang et~al.(2016)Jang, Gu, and Poole}]{jang2016categorical}
\bibinfo{author}{E.~Jang}, \bibinfo{author}{S.~Gu}, \bibinfo{author}{B.~Poole},
\newblock \bibinfo{title}{Categorical reparameterization with gumbel-softmax},
\newblock \bibinfo{journal}{arXiv preprint arXiv:1611.01144}
  (\bibinfo{year}{2016}).
\bibitem[{Todorov et~al.(2012)Todorov, Erez, and Tassa}]{todorov2012mujoco}
\bibinfo{author}{E.~Todorov}, \bibinfo{author}{T.~Erez},
  \bibinfo{author}{Y.~Tassa},
\newblock \bibinfo{title}{Mujoco: A physics engine for model-based control},
\newblock in: \bibinfo{booktitle}{2012 IEEE/RSJ international conference on
  intelligent robots and systems}, \bibinfo{organization}{IEEE},
  \bibinfo{year}{2012}, pp. \bibinfo{pages}{5026--5033}.
\bibitem[{Mujoco(2023)}]{halfcheetah}
\bibinfo{author}{Mujoco}, \bibinfo{title}{Mujoco halfcheetah-v2.},
  \bibinfo{howpublished}{\url{https://gym.openai.com/envs/HalfCheetah-v2/.}},
  \bibinfo{year}{2023}.
\bibitem[{Kumar et~al.(2020)Kumar, Zhou, Tucker, and
  Levine}]{kumar2020conservative}
\bibinfo{author}{A.~Kumar}, \bibinfo{author}{A.~Zhou},
  \bibinfo{author}{G.~Tucker}, \bibinfo{author}{S.~Levine},
\newblock \bibinfo{title}{Conservative q-learning for offline reinforcement
  learning},
\newblock \bibinfo{journal}{Advances in Neural Information Processing Systems}
  \bibinfo{volume}{33} (\bibinfo{year}{2020}) \bibinfo{pages}{1179--1191}.
\bibitem[{Agarap(2018)}]{agarap2018deep}
\bibinfo{author}{A.~F. Agarap},
\newblock \bibinfo{title}{Deep learning using rectified linear units (relu)},
\newblock \bibinfo{journal}{arXiv preprint arXiv:1803.08375}
  (\bibinfo{year}{2018}).
\bibitem[{Dann et~al.(2022)Dann, Mansour, Mohri, Sekhari, and
  Sridharan}]{dann2022guarantees}
\bibinfo{author}{C.~Dann}, \bibinfo{author}{Y.~Mansour},
  \bibinfo{author}{M.~Mohri}, \bibinfo{author}{A.~Sekhari},
  \bibinfo{author}{K.~Sridharan},
\newblock \bibinfo{title}{Guarantees for epsilon-greedy reinforcement learning
  with function approximation},
\newblock in: \bibinfo{booktitle}{International conference on machine
  learning}, \bibinfo{organization}{PMLR}, \bibinfo{year}{2022}, pp.
  \bibinfo{pages}{4666--4689}.
\bibitem[{Fujimoto et~al.(2019)Fujimoto, Meger, and Precup}]{fujimoto2019off}
\bibinfo{author}{S.~Fujimoto}, \bibinfo{author}{D.~Meger},
  \bibinfo{author}{D.~Precup},
\newblock \bibinfo{title}{Off-policy deep reinforcement learning without
  exploration},
\newblock in: \bibinfo{booktitle}{International conference on machine
  learning}, \bibinfo{organization}{PMLR}, \bibinfo{year}{2019}, pp.
  \bibinfo{pages}{2052--2062}.
\bibitem[{Fu et~al.(2020)Fu, Kumar, Nachum, Tucker, and Levine}]{fu2020d4rl}
\bibinfo{author}{J.~Fu}, \bibinfo{author}{A.~Kumar},
  \bibinfo{author}{O.~Nachum}, \bibinfo{author}{G.~Tucker},
  \bibinfo{author}{S.~Levine},
\newblock \bibinfo{title}{D4rl: Datasets for deep data-driven reinforcement
  learning},
\newblock \bibinfo{journal}{arXiv preprint arXiv:2004.07219}
  (\bibinfo{year}{2020}).
\bibitem[{Kudo and Richardson(2018)}]{kudo2018sentencepiece}
\bibinfo{author}{T.~Kudo}, \bibinfo{author}{J.~Richardson},
\newblock \bibinfo{title}{Sentencepiece: A simple and language independent
  subword tokenizer and detokenizer for neural text processing},
\newblock \bibinfo{journal}{arXiv preprint arXiv:1808.06226}
  (\bibinfo{year}{2018}).
\bibitem[{Hochreiter and Schmidhuber(1997)}]{hochreiter1997long}
\bibinfo{author}{S.~Hochreiter}, \bibinfo{author}{J.~Schmidhuber},
\newblock \bibinfo{title}{Long short-term memory},
\newblock \bibinfo{journal}{Neural computation} \bibinfo{volume}{9}
  (\bibinfo{year}{1997}) \bibinfo{pages}{1735--1780}.
\bibitem[{Williamson and Forbes(2014)}]{williamson2014introduction}
\bibinfo{author}{E.~J. Williamson}, \bibinfo{author}{A.~Forbes},
\newblock \bibinfo{title}{Introduction to propensity scores},
\newblock \bibinfo{journal}{Respirology} \bibinfo{volume}{19}
  (\bibinfo{year}{2014}) \bibinfo{pages}{625--635}.
\bibitem[{Juozapaitis et~al.(2019)Juozapaitis, Koul, Fern, Erwig, and
  Doshi-Velez}]{juozapaitis2019explainable}
\bibinfo{author}{Z.~Juozapaitis}, \bibinfo{author}{A.~Koul},
  \bibinfo{author}{A.~Fern}, \bibinfo{author}{M.~Erwig},
  \bibinfo{author}{F.~Doshi-Velez},
\newblock \bibinfo{title}{Explainable reinforcement learning via reward
  decomposition},
\newblock in: \bibinfo{booktitle}{IJCAI/ECAI Workshop on explainable artificial
  intelligence}, \bibinfo{year}{2019}.
\bibitem[{Wang et~al.(2020)Wang, Zhang, Kim, and Gu}]{wang2020shapley}
\bibinfo{author}{J.~Wang}, \bibinfo{author}{Y.~Zhang}, \bibinfo{author}{T.-K.
  Kim}, \bibinfo{author}{Y.~Gu},
\newblock \bibinfo{title}{Shapley q-value: A local reward approach to solve
  global reward games},
\newblock in: \bibinfo{booktitle}{Proceedings of the AAAI Conference on
  Artificial Intelligence}, volume~\bibinfo{volume}{34}, \bibinfo{year}{2020},
  pp. \bibinfo{pages}{7285--7292}.

\end{thebibliography}

\end{document}